\documentclass{article} 
\usepackage{iclr2027_conference,times}
\usepackage{enumitem}

\usepackage{amsmath,amsfonts,bm}

\def\Figref#1{Figure~\ref{#1}}

\def\Secref#1{Section~\ref{#1}}

\def\eqref#1{equation~\ref{#1}}

\def\1{\bm{1}}

\DeclareMathAlphabet{\mathsfit}{\encodingdefault}{\sfdefault}{m}{sl}
\SetMathAlphabet{\mathsfit}{bold}{\encodingdefault}{\sfdefault}{bx}{n}

\newcommand{\E}{\mathbb{E}}

\DeclareMathOperator{\Tr}{Tr}

\usepackage{graphicx}
\usepackage{booktabs}
\usepackage{hyperref}
\usepackage{url}

\usepackage[most]{tcolorbox}
\definecolor{findingbg}{HTML}{EEF3FB}
\definecolor{findingframe}{HTML}{2B5CA8}
\newtcolorbox{finding}{breakable, colback=findingbg, colframe=findingframe,
  boxrule=0.5pt, arc=2pt, left=6pt, right=6pt, top=5pt, bottom=5pt}

\definecolor{xblue}{HTML}{4169E1}
\definecolor{xgreen}{HTML}{036C3A}
\definecolor{xpurple}{HTML}{9838B1}
\definecolor{xslategray}{HTML}{70818F}
\definecolor{xorange}{HTML}{FF8C00}
\definecolor{xcyan}{HTML}{06AEEF}
\definecolor{xred}{HTML}{FF0000}
\definecolor{xgray}{HTML}{808080}
\definecolor{xxgreen}{HTML}{009F86}
\definecolor{xsienna}{HTML}{8B4512}
\definecolor{xxpurple}{HTML}{623E99}

\tcbuselibrary{skins}
\newcommand{\mybox}[2][xblue]{%
  \begin{tcolorbox}[
    enhanced,
    colback=#1!5!white,
    frame hidden,
    boxrule=0pt,
    borderline west={2.5pt}{0pt}{#1},
    sharp corners,
    boxsep=0pt,
    left=8pt, right=8pt,
    top=6pt, bottom=6pt,
    before skip=7pt, after skip=7pt,
    fontupper=\normalsize,
    before upper={\colorlet{boxaccent}{#1}}
  ]
  #2
  \end{tcolorbox}%
}

\usepackage{xcolor}

\definecolor{todobulletcolor}{HTML}{0042a5}

\title{What Does Fr\'echet Distance Measure? \\ A Directional Decomposition}

\author{Yunghee Lee \\
Agency for Defense Development, South Korea\\
\And
Jaeyeon Kim \\
Harvard University \\
}

\iclrfinalcopy 
\begin{document}

\maketitle
\lhead{Preprint}

\begin{abstract}
The Fr\'echet distance is a \emph{de facto} standard for evaluating generative models across domains, appearing as FID for images and FVD for videos. It summarizes the discrepancy between generated and reference distributions in a single scalar, with lower values typically interpreted as better generation quality. However, this scalar view can obscure what drives the comparison. For example, in COCO dataset, increasing the number of diffusion sampling steps improves ImageReward scores yet worsens (increases) FID. Motivated by this mismatch, we seek to make the Fr\'echet distance more interpretable by uncovering \emph{where} the discrepancy lies. To this end, we introduce \emph{directional Fr\'echet distance}, the expected squared projection of the optimal transport displacement onto a given direction.
Across our image, video, and protein case studies, we find that a small number of interpretable directions account for much of the distance. We use these directions to explain the FID increase in terms of semantic concepts represented by CLIP embeddings, quantify FVD's bias toward per-frame appearance, and revisit the interpretation of Protein FID. We open-source our codebase at \url{https://github.com/yhlee-add/directional-fd}.
\end{abstract}

\section{Introduction}
\label{sec:intro}
Fr\'echet distance~\citep{dowson1982frechet} underlies \emph{de facto} evaluation metrics for generative models in continuous domains: Fr\'echet Inception distance (FID) for images~\citep{heusel2017gans}, Fr\'echet video distance (FVD) for videos~\citep{unterthiner2018towards}, and Fr\'echet audio distance (FAD) for audio~\citep{kilgour2018fr}. A lower distance is read as a better generator, so the metric does not merely report progress but also \emph{defines it}, steering architecture, sampling, and hyperparameter choices~\citep{lu2025dpm, kim2025mixdit}. This influential signal, however, remains a \textbf{scalar summary} of a comparison between two high-dimensional feature distributions.

This blind spot surfaces in recurring contradictions. For images, tuning a diffusion sampler configuration to minimize FID can drive per-sample quality to its worst~\citep{lee2026tortoise}; stronger image distortions can even improve (lower) the metric~\citep{jayasumana2024rethinking}; additional sampling steps can produce visibly sharper images~\citep{choi2025enhanced}, yet worsen FID on the COCO dataset~\citep{lin2014microsoft}. For videos, FVD
favors per-frame appearance over motion, so a flickering clip can still score well~\citep{ge2024content}.

Prior work has sought to address these issues by refining Fr\'echet metrics or introducing \emph{scalar alternatives}, such as changing the embedding~\citep{kynkaanniemi2022role}, adopting a kernel two-sample statistic~\citep{binkowski2018demystifying, jayasumana2024rethinking}, or reducing estimator bias~\citep{chong2020effectively}. These changes can mitigate specific failures, but the resulting metrics still do not reveal \emph{where} in feature space the distributions differ or \emph{what} the discrepancy means.

\textbf{Contribution.} In this work, we take a complementary approach: making existing Fr\'echet metrics interpretable. We introduce \textbf{directional Fr\'echet distance}, which measures the contribution of a fixed feature direction to the Fr\'echet distance. These directional distances sum to the Fr\'echet distance over any orthonormal basis, and the top eigenvectors of the displacement second-moment matrix inform where the discrepancy concentrates. We further \emph{affix} these directions to \textbf{interpretable concepts} using CLIP concept bridges~\citep{kim2018interpretability, bhalla2024interpreting} for images and structural-property concept directions for proteins.

This gives \textbf{semantic meaning to directional distances}: a larger distance along a direction associated with a concept, e.g., composition simplicity or brightness, indicates that the metric attributes more of the discrepancy between generated and reference samples to that concept. With this lens, we revisit three established Fr\'echet metrics—FID, FVD, and Protein FID—to explain what drives previously reported discrepancies:
\begin{itemize}[leftmargin=0.8em]
    \item Images: In some reference datasets such as COCO, additional sampling steps can increase ImageReward but also increase FID~\citep{lee2026rethinking}. We find that two-thirds of this increase lies along a single direction in the feature space associated with composition simplicity.
    \item Videos: In the controlled distortions of \citet{ge2024content}, I3D-based FVD responds weakly to added flicker. We find that the flicker band carries less of the distance than the appearance band with I3D features, but more with VideoMAE-v2 features.
    \item Proteins: Protein FID was introduced to detect collapse onto a narrow range of easily foldable structures~\citep{faltings2025protein}, and its observed gap was partly attributed to missing beta sheets. We find that a single direction associated with structural regularity accounts for most of the gap, while the alpha-versus-beta direction accounts for little.
\end{itemize}

\section{Preliminaries}

In Section~\ref{sec:frechet}, we review the Fr\'echet distance and the family of metrics built from it. In Section~\ref{sec:contradictions}, we revisit prior observations across images and videos whose apparent contradictions stem from reading this single number as a measure of per-sample quality.

\subsection{The Fr\'echet distance and its instances}
\label{sec:frechet}

Fr\'echet distance is widely used to evaluate generative models by comparing Gaussian approximations to generated and reference distributions in a fixed feature space. A prominent example is Fr\'echet Inception distance (FID)~\citep{heusel2017gans}, which uses InceptionV3~\citep{szegedy2016rethinking} features. The Fr\'echet distance is the squared 2-Wasserstein distance between the fitted Gaussians, which admits the closed form
\begin{equation}
  \mathrm{FD} = \|\mu_r - \mu_g\|^2 + \Tr \left(\Sigma_r+\Sigma_g-2\left(\Sigma_r^{1/2}\Sigma_g\Sigma_r^{1/2}\right)^{1/2} \right),
  \label{eq:fid_closed_form}
\end{equation} 
where $(\mu_r, \Sigma_r)$ and $(\mu_g, \Sigma_g)$ are the mean and covariance of the reference and generated feature distributions.

The same Fr\'echet distance can also be computed over other embeddings such as CLIP ViT-L/14~\citep{radford2021learning} and DINOv2~\citep{oquab2023dinov2}. The same recipe extends the metric across domains: Video is evaluated with FVD~\citep{unterthiner2018towards}, the Fr\'echet distance over I3D embeddings~\citep{carreira2017quo}, and protein structure generation has recently adopted Protein FID~\citep{faltings2025protein} over ESM3 structure-token embeddings~\citep{hayes2025simulating}.
Across all of these, a full comparison between two distributions is reduced to a single number, which reports \emph{that} they differ but not \emph{where}.

\subsection{Contradictory observations}
\label{sec:contradictions}

A lower Fr\'echet distance is commonly interpreted as a \emph{higher-quality} generator. This interpretation suggests that improvements in per-sample quality should be accompanied by a decrease in distance. In this section, we present two prior observations that challenge this.

To begin with, we first formally define the \emph{per-sample quality}: for a chosen sample-level scoring function $q$, we denote the quality of generator $G$ to be $\mathbb{E}_{x \sim G} [q(x)]$. Examples of such scoring functions include CLIP score~\citep{hessel2021clipscore}, ImageReward~\citep{xu2023imagereward}, and the numerical accuracy of each sample.

\begin{figure}[t]
  \centering
  \includegraphics[width=\linewidth]{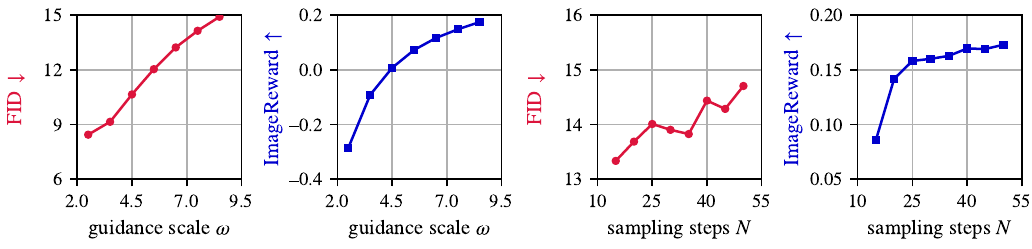}
  \vspace{-22pt}
  \caption{
    \emph{Sign reversal} on SD1.5/COCO under two standard knobs. As the classifier-free guidance scale (left) or the number of sampling steps (right) increases, per-sample quality rises while FID worsens. A lower FID therefore does not track higher per-sample quality.
  }
  \vspace{-10pt}
  \label{fig:sign_reversal}
\end{figure}

\textbf{Images.}
Two standard knobs on Stable Diffusion 1.5~(SD1.5, \citealp{rombach2022high}) make this concrete (Figure~\ref{fig:sign_reversal}). Raising the classifier-free guidance scale~\citep{ho2022classifier} makes each sample match the prompt better and improves ImageReward, yet FID on the COCO reference dataset~\citep{lin2014microsoft} worsens monotonically~\citep[Table 8]{lee2026tortoise}. Using more denoising steps reduces numerical error in each sample and also improves ImageReward~\citep{lee2026rethinking}, but FID can rise~\citep[Table 1]{choi2025enhanced} or converge~\citep[Section 7.2]{lu2025dpm}.

We call this contradictory phenomenon \emph{sign reversal}: a change that raises per-sample quality raises the Fr\'echet distance instead of lowering it.
Sign reversal makes FID an unreliable gate on generation quality. A sampling algorithm~\citep{lu2025dpm} or an inference accelerator~\citep{kim2025mixdit} is accepted once it matches or lowers FID, read as evidence that quality is preserved. Under sign reversal, that inference is invalid, since the same low FID is equally consistent with per-sample quality having dropped, and the verdict on the method no longer follows.

\textbf{Videos.}
FVD computes the Fr\'echet distance in the feature space of I3D~\citep{carreira2017quo}, a video model pretrained for action recognition.
In practice, however, FVD favors per-frame appearance over motion~\citep{ge2024content}: a clip whose individual frames are realistic can still score well when it flickers. To address this, \citet{ge2024content} propose replacing I3D features with self-supervised video embeddings such as VideoMAE-v2~\citep{wang2023videomae}, improving sensitivity to motion while retaining the same Fr\'echet distance computation.

\section{Directional Fr\'echet distance}
\label{sec:method}
The contradictory observations of Section~\ref{sec:contradictions} highlight a limitation of interpreting Fr\'echet distance (FD) as a measure of per-sample quality: FD aggregates discrepancies across feature directions into a single value, obscuring \emph{which} of them drive it.
We therefore introduce \textbf{\emph{directional Fr\'echet distance}}, the contribution of a given direction to FD, which recovers directional information that is encoded in the means and covariances used to compute FD but lost in the final number.

\textbf{Directional Fr\'echet distance.} To define directional Fr\'echet distance, we first express FD as an optimal transport cost. Assuming that two Gaussian distributions $\mathcal{N}(\mu_r, \Sigma_r)$ and $\mathcal{N}(\mu_g, \Sigma_g)$, which are defined by the first and second moments of the reference and generated distributions, are both nondegenerate, the FD in~\eqref{eq:fid_closed_form} is defined by~\citep{knott1984optimal, peyre2019computational}
\begin{equation*}
    \mathrm{FD} = \mathbb{E}[\left\|x - T(x)\right\|^2], \quad T(x):=\mu_g + A(x-\mu_r), \quad  A = \Sigma_r^{-1/2}\left(\Sigma_r^{1/2}\Sigma_g\Sigma_r^{1/2}\right)^{1/2}\Sigma_r^{-1/2}.
\end{equation*}
The expectation is calculated over $x \sim \mathcal{N}(\mu_r, \Sigma_r)$. Here, one can show that $T$ is the optimal transport map between $\mathcal{N}(\mu_r, \Sigma_r)$ and $\mathcal{N}(\mu_g, \Sigma_g)$, i.e., it minimizes the expected squared norm among all measurable maps that transport $\mathcal{N}(\mu_r, \Sigma_r)$ to $\mathcal{N}(\mu_g, \Sigma_g)$:
\begin{equation*}
    T = \mathrm{argmin}_{S} \mathbb{E}_{x\sim \mathcal{N}(\mu_r, \Sigma_r)}[\left\|x-S(x)\right\|^2],
\end{equation*}
where the argmin is taken over all maps $S$ satisfying $S(x)\sim \mathcal{N}(\mu_g, \Sigma_g)$ when $x\sim \mathcal{N}(\mu_r, \Sigma_r)$. 

Now we define \emph{directional Fr\'echet distance}. For a given unit direction $w$, the directional Fr\'echet distance $\mathrm{FD}(w)$ is computed as the squared distance \emph{along} $w$ under the optimal transport map $T$, i.e.,
\begin{equation*}
    \mathrm{FD}(w):=\mathbb{E}_{x \sim \mathcal{N}(\mu_r, \Sigma_r)}[(w^\top  (x - T(x)))^2].
\end{equation*}
An alternative way to formulate $\mathrm{FD}(w)$ is to introduce the \emph{displacement second-moment matrix}~\citep{paty2019subspace, meng2020sufficient} $M = \E_{x\sim\mathcal{N}(\mu_r,\Sigma_r)} \left[(x-T(x))(x-T(x))^\top\right]$, resulting in $\mathrm{FD}(w) = w^\top M w$. Therefore, for any orthonormal basis $\{w_k\}$, directional Fr\'echet distances sum to the Fr\'echet distance, $\mathrm{FD} = \sum_k \mathrm{FD}(w_k)$ (Appendix~\ref{app:methods}).

\textbf{Interpretation.}
A large $\mathrm{FD}(w)$ indicates that displacement along $w$ contributes substantially to the optimal transport cost between the generated and reference Gaussians.
This decomposition thus provides a \emph{diagnostic tool} for understanding what drives FD.
For instance, when a direction is associated with an interpretable concept, such as brightness or composition simplicity, its directional FD quantifies how much of the discrepancy between the generated and reference distributions is attributed to that concept direction.

\section{Experiments}
\label{sec:experiments}

In this section, we use directional Fr\'echet distance to revisit three Fr\'echet metrics.
We first explain why additional denoising steps raise ImageReward but also raise FID, and why the sign of this effect flips across references (\Secref{sec:sign_reversal}).
We then explain why VideoMAE-v2 makes FVD more sensitive to flicker than I3D (\Secref{sec:video}).
Finally, we examine which structural directions drive the Protein FID gap~\citep{faltings2025protein} (\Secref{sec:protein}).

Each case study follows a common procedure.
We take the top eigenvectors $e_1,e_2,\dots$ of a displacement second-moment matrix $M$ (\Secref{sec:method}), which identify the directions along which a distance concentrates.
We then interpret these directions within each domain (Appendix~\ref{app:experiments}).

\subsection{Sign reversal}
\label{sec:sign_reversal}

We study sign reversal on images in the common setting of FID on the COCO reference, where additional denoising steps on Stable Diffusion 1.5 (SD1.5) raise ImageReward score but also raise FID (\Figref{fig:sign_reversal}).
We build the explanation in two steps: (1) what the top directions mean, (2) then which of them carries the change in FID and why its sign flips across references.

We first analyze a single comparison, SD1.5 at $50$ steps against COCO, to find the directions FID concentrates on and name them with concept dictionaries (\Secref{sec:sign_reversal_one}).
We then sweep the denoising steps and follow which of these directions carries the change in FID, first on COCO and then across six references, to explain what sets its sign (\Secref{sec:sign_reversal_sweep}).

\subsubsection{Naming the top directions}
\label{sec:sign_reversal_one}

We evaluate SD1.5 at $50$ denoising steps against the COCO reference in Inception feature space, where FID is $14.6$, and ask three questions of this single comparison: where in feature space the distance concentrates, what those directions mean, and how the two distributions differ along them.

\textbf{Where: FID concentrates on a few directions.}
Ordering the eigenvectors $e_1, e_2, \dots$ of $M$ by their directional FD ranks the feature directions by the share of FID they carry, and the spectrum is steep: $e_1$ carries $24.0\%$ of FID, $e_2$ carries $3.8\%$, and $e_3$ carries $2.5\%$, so three eigenvectors account for $30.3\%$ of a distance computed across $2048$ feature dimensions.

\textbf{What: CLIP as a bridge.}
A substantial fraction of FID thus concentrates on three directions, but these are directions in Inception feature space and have no semantics on their own, so naming them is what turns the decomposition into a statement about how the reference and generated sets differ.
We attach meaning with CLIP as a bridge between text and the Inception features FID is computed in.
We write a CLIP dictionary of $24$ photographic concepts, each a pair of opposing prompt sets, e.g. a single clean subject against a cluttered scene, and embed both poles with the CLIP text encoder.

For each concept, we rank the reference images by CLIP similarity to its text direction and take the top and bottom quantiles as the aligned and anti-aligned groups, and the difference of their mean Inception features is the concept direction $w_c$, a supervised direction in the same space as the eigenvectors, similar to the concept-activation construction of \citet{kim2018interpretability}.
Given the concept directions, we fit each eigenvector as a sparse signed combination of them with a Lasso and take the surviving concepts as the direction's name, as in \citet{bhalla2024interpreting}.
Table~\ref{tab:lasso} lists the result: $e_1$ reconstructs with cosine similarity $0.74$ and is dominated by composition simplicity, so we name it after that concept; it runs from cluttered multi-object scenes toward simple single-subject images.
$e_2$ reconstructs more weakly at $0.49$ as a mixture of brightness, weather, and indoor-outdoor content.

\begin{table}[t]
  \centering
  \vspace{-5pt}
  \caption{
    Sparse concept fit of the top eigenvectors of $M$ over the two concept dictionaries, for SD1.5, $50$ steps, and COCO. Share is $\mathrm{FD}(e_i)/\mathrm{FD}$, the weights are the largest surviving Lasso coefficients, and cos is the reconstruction cosine.
  }
  \label{tab:lasso}
  \small
  \vspace{5pt}
  \begin{tabular}{@{}cllp{9cm}c@{}}
    \toprule
    & share & dictionary & surviving concepts (weight) & cos \\
    \midrule
    $e_1$ & $24.0\%$ & CLIP & composition simplicity ($0.55$), depth of field ($0.12$), furniture ($0.08$) & $0.74$ \\
    $e_2$ & $3.8\%$ & CLIP & brightness ($-0.15$), weather ($-0.12$), crowd density ($0.08$), indoor/outdoor ($0.08$) & $0.49$ \\
    $e_3$ & $2.5\%$ & ImageNet & Dungeness crab ($-0.09$), mashed potato ($-0.06$), burrito ($-0.05$) & $0.40$ \\
    \bottomrule
  \end{tabular}
  \vspace{-10pt}
\end{table}

\textbf{What: ImageNet classes name what CLIP concepts miss.}
The CLIP dictionary does not reach $e_3$, which reconstructs to cosine similarity $0.17$ with no dominant concept, so we turn to a second dictionary built without prompts.
The Inception features map to the $1008$ ImageNet logits through one affine layer, so the unit-normalized rows of that layer form a data-free concept dictionary~\citep{kynkaanniemi2022role}.
A sparse fit over these class rows names $e_3$ as a consistent food cluster (Dungeness crab, mashed potato, burrito, ice cream), so together the two dictionaries name the top three eigenvectors of $M$.

\begin{figure}[t]
  \centering
  \includegraphics[width=\linewidth]{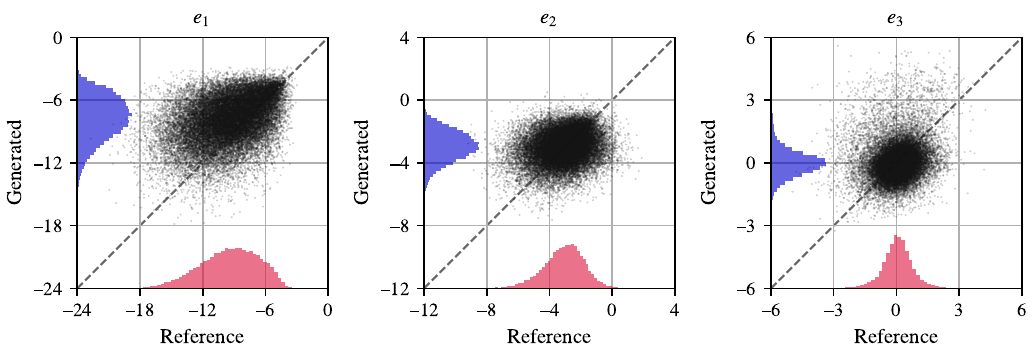}
  \vspace{-22pt}
  \caption{
    Reference and generated features projected onto the top three eigenvectors of $M$, using SD1.5, $50$ steps, and COCO. Each point pairs a caption's reference and generated projection. The reference (red) and generated (blue) marginal distributions are drawn inside, and the dashed line is $y=x$. For $e_1$, the points are slightly above $y=x$, indicating that the generated set is shifted toward $+e_1$. For $e_2$, the points concentrate vertically, demonstrating that the generated set has less variance, and vice versa for $e_3$.
  }
  \vspace{-10pt}
  \label{fig:sign_reversal_proj}
\end{figure}

\textbf{How: a shift in mean or a change in spread.}
Projecting the two sets onto each $e_i$ (\Figref{fig:sign_reversal_proj}) tells whether they differ in mean or in spread along it.
On $e_1$, the generated mean sits toward the simple-composition end (Cohen's $d = 0.69$), so the generated set is shifted toward simpler, more single-object images than COCO.
On $e_2$ and $e_3$, the means coincide and only the spread differs: the generated set has $0.56$ times the reference variance along $e_2$, holding a narrower range of lighting and indoor-outdoor content, and $1.16$ times along $e_3$, spreading more widely over food content.

\mybox{
  \textbf{Finding.} The direction that carries the most directional FD, $e_1$, is dominated by composition simplicity, and the generated set differs from COCO by a shift along it, toward simpler, single-object images.
}

\textbf{Robustness check: DINOv2.} To test whether this finding holds across feature spaces, we repeat the decomposition in DINOv2 feature space~\citep{oquab2023dinov2}. The dominant direction has a similar interpretation: $e_1$ captures composition simplicity, accounts for $14.0\%$ of the distance, and reconstructs at a cosine similarity of $0.41$. The generated set is again shifted relative to COCO toward simpler compositions (Cohen's $d=2.2$). Thus, the interpretation of the dominant direction is consistent across embeddings.

\subsubsection{Which directions carry the reversal}
\label{sec:sign_reversal_sweep}
We next use this decomposition to identify which directions account for sign reversal. Using SD1.5 and COCO again, we take eight generated sets using $\{15, 20, \dots, 50\}$ steps. 

\begin{figure}[t]
  \centering
  \includegraphics[width=\linewidth]{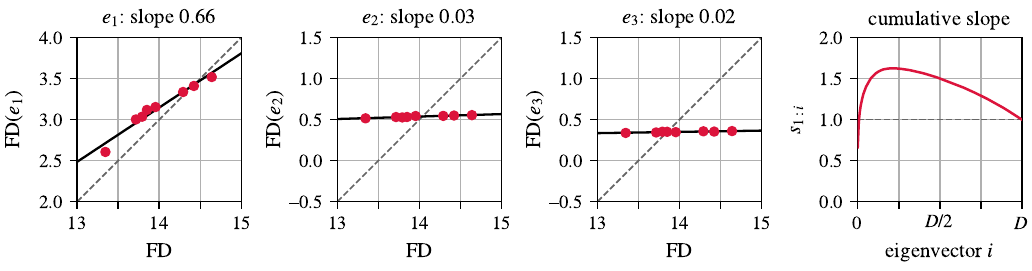}
  \vspace{-22pt}
  \caption{
    \emph{Left}: directional FD of the top three fixed eigenvectors against total FID across the eight-step COCO sweep, in the frozen $50$-step basis. Each point is one step setting, the solid line is the least-squares fit, and its slope is the fraction of the FID change that eigenvector carries, and the dashed line is the $\mathrm{slope}=1$ an eigenvector carrying the entire change would follow. The top eigenvector $e_1$ carries most of it (slope $0.66$, $R^2 = 0.96$), while $e_2$ and $e_3$ stay nearly flat (slopes $0.03$ and $0.02$).
    \emph{Right}: cumulative slope $s_{1:i}$, the summed slopes of the top $i$ eigenvectors, which rises above the whole change (dashed) over the head and falls back to it over the tail (the eigenvectors before and after the curve's peak).
  }
  \vspace{-10pt}
  \label{fig:reversal_shares}
\end{figure}

To track the same directions across these settings, we fix the eigenbasis of the $50$-step $M$ and regress each direction's $\mathrm{FD}(e_i)$ against the total FID; since the directional FDs over the basis sum to FD, the slopes sum to one, and each slope is the fraction of the FD change the direction carries (\Figref{fig:reversal_shares}).

\textbf{One eigenvector accounts for two thirds of the increase.}
The top eigenvector $e_1$ has slope $0.66$ ($R^2 = 0.96$), while $e_2$ and $e_3$ stay nearly flat at slopes $0.03$ and $0.02$, so two thirds of the FID increase lives on $e_1$ alone.
FID thus rises with steps mainly because the generated mean drifts further toward simpler, single-object images and away from COCO, which ties most of the sign reversal of \Secref{sec:contradictions} to one named direction (Appendix~\ref{app:image}).

\textbf{Rise over the head, fall over the tail.}
The slope regression says how much each eigenvector carries but not where the change accumulates along the spectrum.
To see that, we sum the slopes of the top $i$ eigenvectors into the cumulative slope $s_{1:i}$ and plot it against $i$ (\Figref{fig:reversal_shares}, right).
Since the directional FDs of all $D$ eigenvectors sum to FID, $s_{1:D} = 1$.
The curve rises above one, peaking at $1.62$ near $i = 440$, and then falls back to one; we call the eigenvectors up to this peak the \emph{head} and the rest the \emph{tail}.
The rise opens with the jump of $e_1$, along which the generated mean drifts further toward simpler images as steps increase (Cohen's $d$ grows from $+0.58$ to $+0.69$), and continues with small gains over the other head eigenvectors.
Over the tail the curve falls, as more steps contract the generated variance there by about $1.5\%$ from $15$ to $50$ steps, lowering directional FD by small amounts across the tail eigenvectors.
On COCO the rise outweighs the fall, so FID rises overall (sign reversal).
Appendix~\ref{app:reversal_cumulative} repeats this sweep on six references.

\mybox{
  \textbf{Finding.} Two thirds of the step-driven FID increase on COCO lies on a single eigenvector, $e_1$, named after composition simplicity, along which the generated mean drifts toward simpler images. Over the whole spectrum, the rise over the head outweighs a broad fall over the tail.
}

\subsection{Video content bias}
\label{sec:video}

Fr\'echet video distance~(FVD,~\citealp{unterthiner2018towards}) is the Fr\'echet distance over the features of I3D~\citep{carreira2017quo}, which is an action-recognition model.
\citet{ge2024content} expose the \emph{content bias} of FVD (\Secref{sec:contradictions}) by decoupling per-frame appearance from motion. In particular, they create two sets of degraded real videos: one with the same distortion across frames and another with a distortion of matched severity resampled each frame. They find that I3D-based FVD barely penalizes the resulting flicker.

To address this bias, they replace I3D with VideoMAE-v2~\citep{wang2023videomae}, which makes FVD more sensitive to flicker. We use directional FD to explain this difference: in I3D feature space, appearance-related directions account for more of the spatio-temporally distorted set's distance from real videos than flicker-related directions, whereas in VideoMAE-v2 feature space, the ordering reverses.

\textbf{Decoupled distortions as probes.}
We obtain these directions from the two degraded sets of \citet{ge2024content}, which we rebuild on a reference set $R$ of $2048$ UCF-101 clips of $16$ frames~\citep{soomro2012ucf101}.
The distortions are elastic transformation and motion blur, each at the five severities of ImageNet-C~\citep{hendrycks2019benchmarking}.
The \emph{appearance set} $S$ is their spatially distorted set, which applies one sampled distortion to every frame of a clip.
The \emph{flicker set} $T$ is their spatio-temporally distorted set, which resamples the distortion at every frame.
The two sets share the distribution of per-frame distortions, so $T$ differs from $S$ only by the flicker.
%

\textbf{Appearance and flicker displacements.}
We write $M(P, Q)$ for the displacement second-moment matrix between two sets $P$ and $Q$.
The \emph{appearance displacement} $M(R, S)$ reflects the distortion itself.
The \emph{flicker displacement} $M(S, T)$ reflects the flicker added on top of it.
We average each displacement over severities and take its top eigenvectors as the \emph{appearance band} and the \emph{flicker band}.
The cumulative share $\sum_{i \le k} \mathrm{FD}(e_i) / \mathrm{FD}$ is the fraction of a set's FD that the top $k$ eigenvectors $e_i$ of a band carry.
The two bands are computed independently and can overlap.
A share therefore measures how much of a distance aligns with a band, not an exclusive allocation between the bands (Appendix~\ref{app:video}).

\subsubsection{Which band carries the flicker set's distance}
\label{sec:video_flicker}

We evaluate the displacement $M(R, T)$ between the reference and the flicker set on both bands.
The share on the appearance band shows how much of this distance aligns with the distortion itself.
The share on the flicker band shows how much aligns with the flicker.

\textbf{The two embeddings put the flicker set's distance on opposite bands.}
Table~\ref{tab:video_bands} gives both shares at severity $3$, with bands averaged over the other four severities.
I3D puts far more of this distance on the appearance band than on the flicker band: with motion blur, $0.33$ against $0.02$ on the top eigenvector and $0.66$ against $0.27$ at $k = 10$.
I3D thus sees the flicker set mostly through the distortion it shares with the appearance set.
VideoMAE reverses the ordering, with $0.06$ against $0.47$ on the top eigenvector and $0.18$ against $0.61$ at $k = 10$.
Both distortion families show this flip, and motion blur shows it more clearly.
The flip with motion blur also holds when either band is orthogonalized against the other (Appendix~\ref{app:video}).

\begin{table}[t]
  \centering
  \vspace{-5pt}
  \caption{
    Share of directional FD on the appearance and flicker bands (the top eigenvectors of $M(R, S)$ and $M(S, T)$), at the top $k = 1$ and $k = 10$ eigenvectors.
    The flicker-set panel evaluates $M(R, T)$ at severity $3$, with bands averaged over the other four severities.
    The generator panel evaluates ModelScope's generated set, zero-shot on the UCF-101 class names, on bands averaged over all five severities.
    The two bands overlap, so the shares of a row can sum past one.
  }
  \label{tab:video_bands}
  \small
  \vspace{5pt}
  \setlength{\tabcolsep}{3.4pt}
  \begin{tabular}{@{}lcccccccc@{}}
    \toprule
    & \multicolumn{4}{c}{flicker set, $M(R, T)$} & \multicolumn{4}{c}{generator, $M(R, G)$} \\
    \cmidrule(lr){2-5}\cmidrule(lr){6-9}
    & \multicolumn{2}{c}{$k = 1$} & \multicolumn{2}{c}{$k = 10$} & \multicolumn{2}{c}{$k = 1$} & \multicolumn{2}{c}{$k = 10$} \\
    \cmidrule(lr){2-3}\cmidrule(lr){4-5}\cmidrule(lr){6-7}\cmidrule(lr){8-9}
    embedding, family & appearance & flicker & appearance & flicker & appearance & flicker & appearance & flicker \\
    \midrule
    I3D, motion blur & $0.325$ & $0.023$ & $0.663$ & $0.273$ & $0.022$ & $0.005$ & $0.288$ & $0.150$ \\
    I3D, elastic & $0.372$ & $0.211$ & $0.648$ & $0.495$ & $0.027$ & $0.012$ & $0.288$ & $0.107$ \\
    VideoMAE, motion blur & $0.061$ & $0.470$ & $0.181$ & $0.612$ & $0.013$ & $0.019$ & $0.095$ & $0.096$ \\
    VideoMAE, elastic & $0.163$ & $0.389$ & $0.282$ & $0.553$ & $0.012$ & $0.009$ & $0.148$ & $0.084$ \\
    \bottomrule
  \end{tabular}
  \vspace{-10pt}
\end{table}

\textbf{The band share holds where the scalar ratio fades.}
\citet{ge2024content} measure the bias with the ratio of the flicker set's FVD to the appearance set's.
This ratio shrinks toward $1$ as severity grows, because the appearance set's FVD grows faster than what the flicker adds.
I3D's ratio with motion blur falls from $1.41$ at the mildest severity to $1.07$ at the harshest.
The band share is a different quantity, and it keeps the two embeddings apart at every severity.
VideoMAE's share on the top motion blur flicker eigenvector stays above $0.23$, while I3D's stays under $0.06$ (Appendix~\ref{app:video}).

\mybox{
  \textbf{Finding.} I3D puts more of the flicker set's distance on the appearance band, the directions the distortion already moves, than on the flicker band. VideoMAE does the reverse. The content bias of FVD is this small flicker-band share under I3D.
}

\subsubsection{A real generator on the bands}

The bands so far come from controlled distortions.
We now ask whether the content bias found on the flicker set (\Secref{sec:video_flicker}) also appears in a real generator's discrepancy from the reference.
We sample a set $G$ of $2048$ clips with ModelScope text-to-video~\citep{wang2023modelscope}, zero-shot on the UCF-101 class names.
We then evaluate its displacement $M(R, G)$ on the appearance and flicker bands under both embeddings.
%

\textbf{VideoMAE shifts the generated distance toward the flicker band.}
I3D puts about twice as much of $M(R, G)$ on the appearance band as on the flicker band, with $0.29$ against $0.15$ on the motion blur bands at $k = 10$ (Table~\ref{tab:video_bands}).
This is the same ordering as on the flicker set.
VideoMAE splits the same distance evenly, with $0.095$ against $0.096$.
The ratio of the flicker share to the appearance share thus rises from $0.52$ under I3D to $1.01$ under VideoMAE.
The elastic bands show the same shift, from $0.37$ to $0.57$, although appearance still leads there (Appendix~\ref{app:video}).

\mybox{
  \textbf{Finding.} VideoMAE shifts a real generator's distance toward the flicker band. The two motion blur bands carry it evenly under VideoMAE, while I3D puts twice as much on the appearance band. A scalar ratio cannot report this shift upon an arbitrary generated set.
}

\subsection{Protein FID attribution}
\label{sec:protein}

Designability is the most widely used per-sample metric for protein backbones: An inverse-folding model designs a sequence for the backbone, which is then refolded by a structure predictor. The backbone is considered designable if the predicted structure matches it within a self-consistency RMSD (scRMSD) of $2$\,\AA~\citep{trippe2022diffusion,watson2023novo}.

High designability, however, can be achieved by generating mostly alpha-helical backbones: alpha helices rely largely on local hydrogen bonds, whereas beta sheets require long-range coordination. Optimizing for designability can therefore favor a narrow range of easily foldable, helix-rich structures while losing beta sheets from the generated distribution.

To detect such collapse, \citet{faltings2025protein} introduce Protein FID, the Fr\'echet distance between generated and reference structures embedded with the ESM3 structure encoder~\citep{hayes2025simulating}. They validate it through its monotone response to structural perturbations and its recovery of the FoldSeek and CATH hierarchies.

The authors partly attribute the Protein FID gap to missing beta sheets, based on a sample-level influence analysis that identifies large alpha-helical bundles as major contributors. We use directional FD to examine which structural directions drive the gap: the alpha-versus-beta contrast accounts for only a small share, whereas a single direction associated with structural regularity accounts for the largest share.

\textbf{Reference and embedding.}
Protein FID mean-pools the per-residue ESM3 embeddings to $1536$ dimensions.
It then applies a PCA to $32$ dimensions, fit jointly over the reference and generated embeddings, because the raw covariances are nearly degenerate~\citep{faltings2025protein}.
We analyze this $32$-dimensional space and recompute the optimal transport in it.
The reference is the released embeddings of curated PDB structures.
The raw space gives the same picture (Appendix~\ref{app:protein}).

\textbf{Generations and concept directions.}
We generate unconditional backbones with La-Proteina~\citep{geffner2026proteina}, $467$ structures per setting, following the Protein FID protocol.
To interpret the eigenvectors of $M$, we build a concept dictionary in the same space. We define the alpha-versus-beta direction using the mainly-alpha and mainly-beta CATH classes~\citep{sillitoe2021cath}, and additional concepts by splitting ten whole-structure geometric properties into terciles. Each concept direction is a difference between group means of CATH embeddings, projected onto the same PCA basis. We then fit each eigenvector with a Lasso over this dictionary, as in \Secref{sec:sign_reversal_one}.

\subsubsection{Naming the dominant direction}
\label{sec:protein_naming}

We analyze La-Proteina at $400$ steps against the reference and ask the three questions of \Secref{sec:sign_reversal_one}: where, what, and how.

\textbf{Where: the distance concentrates on one direction.}
The spectrum of $M$ is steep: $e_1$ carries $66\%$ of Protein FID, $e_2$ carries $14\%$, and $e_3$ carries $4\%$.
The top direction is not an arbitrary shift of the generator away from the reference.
The reference itself varies along $e_1$ more than along $94\%$ of its principal directions, so $e_1$ is a significant direction of the reference data.

\textbf{What: structural regularity.}
Table~\ref{tab:protein} fits the top eigenvectors over the concept dictionary.
$e_1$ reconstructs at cosine $0.72$ and is dominated by a coil deficit.
The concept directions most aligned with $e_1$ on their own are coil ($-0.46$), helix ($+0.32$), and clash ($-0.29$).
The generated end of $e_1$ is thus the regular, idealized one, and we name $e_1$ the \emph{structural-regularity direction}.
The lower eigenvectors read as backbone-bond uniformity and clash ($e_2$) and as the contact-order residual ($e_3$).

\textbf{What: alpha versus beta is a minor part.}
The alpha-versus-beta direction carries only $5.5\%$ of Protein FID and aligns weakly with the mean shift (cosine $0.24$).
It stays minor even though the generated set is extreme in composition: $71\%$ helix and $5\%$ strand, with $58\%$ of samples containing no beta sheet at all.
Protein FID charges these helix-rich samples mainly along the structural-regularity direction, and only a little along the alpha-versus-beta direction.

\textbf{How: a shift in mean or a change in spread.}
Projecting the two sets onto each $e_i$ tells whether they differ in mean or in spread along it.
The generated mean sits toward the regular end of $e_1$ (Cohen's $d = 3.2$), a shift onto over-idealized backbones.
The means coincide along $e_2$ and $e_3$, and only the spread differs.
The generated set holds only $0.05$ and $0.16$ times the reference variance there.
Its samples are nearly identical in backbone uniformity and contact order, where real proteins vary.

\begin{table}[t]
  \centering
  \vspace{-5pt}
  \caption{
    Sparse concept fit of the top eigenvectors of $M$ over the geometry concept dictionary, and how the two sets differ along each, for La-Proteina at $400$ steps against the ESM3 reference, in the $32$-dimensional PCA space of Protein FID. Share is $\mathrm{FD}(e_i)/\mathrm{FD}$, the weights are Lasso coefficients, cos is the reconstruction cosine, $d$ is the Cohen's $d$ of the reference and generated projections onto $e_i$, and var is their variance ratio. We show the top surviving concepts here. Refer to Appendix~\ref{app:protein} for details.
  }
  \label{tab:protein}
  \small
  \vspace{5pt}
  \begin{tabular}{@{}clp{6.2cm}ccc@{}}
    \toprule
    & share & surviving concepts (weight) & cos & $d$ & var \\
    \midrule
    $e_1$ & $66\%$ & coil ($-0.97$), contact order ($+0.53$), length ($+0.16$), globularity ($-0.16$) & $0.72$ & $+3.2$ & $0.55$ \\
    $e_2$ & $14\%$ & backbone-bond deviation ($-0.58$), clash ($+0.50$) & $0.92$ & $+0.0$ & $0.05$ \\
    $e_3$ & $4\%$ & contact order ($-0.90$), clash ($+0.31$) & $0.83$ & $+0.1$ & $0.16$ \\
    \bottomrule
  \end{tabular}
  \vspace{-10pt}
\end{table}

\mybox{
  \textbf{Finding.} The dominant Protein FID direction, $e_1$, is the structural-regularity direction. It carries two thirds of Protein FID, and the generated set is shifted along it toward low-coil, compact, idealized backbones. The alpha-versus-beta direction carries only $5.5\%$, so Protein FID charges over-regularity far more than the missing beta sheets.
}

\section{Conclusion}
In this work, we introduced directional Fr\'echet distance, decomposing Fr\'echet distance along an orthonormal basis of feature directions. Across images, videos, and proteins, this viewpoint facilitates explaining the sign reversal of FID and the content bias of FVD, and revisits a prior interpretation of Protein FID.

\textbf{Outlook.} More broadly, we hope this work prompts a rethinking of what scalar evaluation can and cannot tell us. A scalar score is useful \emph{per se}--it makes models easy to compare, spurring competition and progress in generative modeling. Nevertheless, it does not reveal which aspects of a distribution a model fails to capture. Directional FD opens the door to more interpretable evaluation of generative models.

\newpage

\subsubsection*{Acknowledgments}
This work was supported by the Agency For Defense Development Grant Funded by the Korean Government (912A45701).

\bibliography{iclr2027_conference}
\bibliographystyle{iclr2027_conference}

\clearpage
\appendix

\section{Related Work}
\label{sec:related}

\paragraph{Critiques of FID.}
A line of work critiques the estimator and the space it lives in.
Inception embeddings of standard datasets fail multivariate normality tests, violating the Gaussian assumption~\citep{jayasumana2024rethinking}.
FID is a biased estimator whose bias depends on the model being evaluated, not just on the sample count~\citep{chong2020effectively}.
Differences in image resizing also shift it by several points on identical data~\citep{parmar2022aliased}.
The representation matters as much as the estimator.
The second-last layer of InceptionV3 is one affine transform from the ImageNet logits, so FID is sensitive to ImageNet class information that need not reflect perceptual quality~\citep{kynkaanniemi2022role}.
The semantics an embedding encodes also depend strongly on its training~\citep{stein2023exposing}.
Surveys of evaluation practice add that the choice of metric is rarely motivated and that automatic metrics often disagree with human judgment~\citep{borji2022pros, otani2023toward}.

\paragraph{Swapping one scalar metric for another.}
Each pathology has typically been mitigated by replacing one scalar metric with another.
Examples include a cleaner embedding~\citep{kynkaanniemi2022role}, a different kernel such as the MMD tests of KID~\citep{binkowski2018demystifying} and CMMD~\citep{jayasumana2024rethinking}, a debiased estimator~\citep{chong2020effectively}, and a content-debiased FVD that traces the video bias to the embedding~\citep{ge2024content}.
Protein FID~\citep{faltings2025protein} is the same move in a new domain, swapping the per-sample designability score for a distributional metric.
These works support their diagnoses with further analyses.
The metric each proposes, however, still reports a single number, which does not show where in feature space the discrepancy lives.

\paragraph{Reporting more than one number.}
The closest attempts to report more than one number measure a generator along two fixed criteria.
Precision and recall separate sample fidelity from distributional coverage~\citep{sajjadi2018assessing, kynkaanniemi2019improved}, and density and coverage make those two criteria more robust to outliers~\citep{naeem2020reliable}.
These measures are informative but preset.
They score a generator on fidelity and diversity, not along data-driven or semantically named directions.

\paragraph{The displacement second-moment matrix.}
The matrix $M$ whose directions we name is not new to optimal transport.
\citet{paty2019subspace} define the second moment $V_\pi = \int (x - y)(x - y)^\top d\pi(x, y)$ of the displacements under a coupling $\pi$ and note that its trace is the transport cost.
They build subspace-robust Wasserstein distances from its top eigenvalues, optimizing the coupling and the subspace jointly.
\citet{meng2020sufficient} use the eigenvectors of the same second moment as a supervised projection for dimension reduction.
Both use the matrix to define a new distance or a new projection.
We instead keep the coupling fixed at the Gaussian optimal map behind an existing metric.
$\Tr M = \mathrm{FD}$ then makes each direction an exact additive share of the Fr\'echet distance that an evaluation already reports.

\paragraph{Interpreting feature directions.}
A separate literature interprets the directions of a feature space rather than the metric.
Concept activation vectors probe a representation along human-named concept directions~\citep{kim2018interpretability}, SpLiCE decomposes CLIP embeddings into a sparse sum of interpretable concepts~\citep{bhalla2024interpreting}, and sparse autoencoders recover interpretable features from protein language models~\citep{simon2025interplm}.
We reuse this machinery, but to name the directions a Fr\'echet metric charges for rather than to interpret an embedding in isolation.

\section{Methodological details}
\label{app:methods}

\paragraph{Recovering the closed form.}
Write $\delta = \mu_r - \mu_g$ and $C = (\Sigma_r^{1/2}\Sigma_g\Sigma_r^{1/2})^{1/2}$.
The map $T$ of \Secref{sec:method} then has $A = \Sigma_r^{-1/2} C \Sigma_r^{-1/2}$, a symmetric matrix with $A\Sigma_r A = \Sigma_g$ and $\Tr(A\Sigma_r) = \Tr C$.
The displacement is $x - T(x) = \delta + (I - A)(x - \mu_r)$.
The term $x - \mu_r$ has zero mean and covariance $\Sigma_r$, so the cross terms vanish in expectation and
\begin{equation}
  M = \delta\delta^\top + (I - A)\Sigma_r(I - A),
  \qquad
  \mathrm{FD}(w) = (w^\top\delta)^2 + w^\top(I - A)\Sigma_r(I - A)\,w.
  \label{eq:mean_cov}
\end{equation}
Taking the trace recovers the closed form~\eqref{eq:fid_closed_form},
\begin{equation*}
  \Tr M = \|\delta\|^2 + \Tr\Sigma_r - 2\Tr(A\Sigma_r) + \Tr(A\Sigma_r A) = \|\mu_r - \mu_g\|^2 + \Tr\left(\Sigma_r + \Sigma_g - 2C\right).
\end{equation*}
We call the first term of $\mathrm{FD}(w)$ in~\eqref{eq:mean_cov} its \emph{mean part}, the mean gap read along $w$.
We call the second term its \emph{covariance part}, what the map pays along $w$ to carry the reference spread onto the generated one.
Summed over an orthonormal basis, the two parts give the two terms of the closed form.
The mean part of $M$ has rank one, so a dominant mean gap draws the top eigenvector $e_1$ toward the mean-difference direction $\delta / \|\delta\|$.
The covariance part then decides the eigenvectors that follow.

\paragraph{Directional FD is not the FD of a projection.}
The pair $(w^\top x, w^\top T(x))$ with $x \sim \mathcal{N}(\mu_r, \Sigma_r)$ is one coupling of the two projected Gaussians.
Its cost $\mathrm{FD}(w)$ is therefore at least their optimal cost, the Fr\'echet distance $(w^\top\delta)^2 + (\sqrt{w^\top\Sigma_r w} - \sqrt{w^\top\Sigma_g w})^2$ between the projections.
The inequality can be strict, because the optimal coupling of the full distributions need not couple the projections optimally.
For example, two covariances may differ only in how their coordinates correlate.
Their projections onto each coordinate then coincide, but the map still pays along every coordinate to rotate the correlation.
The projected means and variances, which we report as Cohen's $d$ and the variance ratio, thus describe how the two sets differ along $w$.
The same bound holds for a subspace.
The directional FD of a subspace with orthogonal projector $P$ is $\Tr(PM)$, the sum of $\mathrm{FD}(w)$ over an orthonormal basis of the subspace.
Protein FID is the Fr\'echet distance between the projections onto a $32$-dimensional PCA subspace.
It is therefore not the directional FD of that subspace in the raw space, but a lower bound on it.

\paragraph{Singular covariances.}
\Secref{sec:method} assumes both Gaussians nondegenerate, but the map needs only a nondegenerate $\Sigma_r$.
$A$ is then symmetric positive semidefinite for any positive semidefinite $\Sigma_g$, with $A\Sigma_r A = \Sigma_g$.
$T$ thus carries $\mathcal{N}(\mu_r, \Sigma_r)$ to $\mathcal{N}(\mu_g, \Sigma_g)$.
$T$ is also the gradient of a convex function, so by Brenier's theorem~\citep{brenier1991polar} it is the unique optimal map from the absolutely continuous source.
This covers the raw $1536$-dimensional protein analysis of Appendix~\ref{app:protein}.
There the covariance of $467$ generated structures has rank at most $466$, while that of the $4943$ reference structures is nondegenerate.
Both covariances are nondegenerate in the $32$-dimensional PCA space of Protein FID and in every image and video comparison.

\paragraph{Numerical computation.}
We compute in float64 and take every matrix square root from an eigendecomposition, which is real for any positive semidefinite input.
Most implementations of FD instead apply \texttt{scipy.linalg.sqrtm}, a Schur decomposition, to the non-symmetric product of the two covariances.
This returns spurious imaginary parts when the product is singular, although the true square root is real.
The reference implementation of Protein FID~\citep{faltings2025protein} is one of them and rejects the result once the imaginary part exceeds a fixed tolerance.
It therefore fails on a generated set of $32$ structures, whose covariance is singular in the $32$-dimensional PCA space, and succeeds on $40$.
Directional FD also needs the inverse square root in $A$.
We avoid it by assembling the covariance part of $M$ as $\Sigma_r + \Sigma_g - K - K^\top$, where $K = \E[(x - \mu_r)(T(x) - \mu_g)^\top] = \Sigma_r A$ is the cross-covariance under the optimal coupling.
We obtain $K = \Sigma_r^{1/2} W^\top \Sigma_g^{1/2}$ from the polar decomposition $\Sigma_g^{1/2}\Sigma_r^{1/2} = WP$ into an orthogonal $W$ and a positive semidefinite $P$.
This polar decomposition requires no inversion and exists for singular covariances too.

\section{Experimental details}
\label{app:experiments}

\paragraph{Shared analysis.}
Each eigenvector $e_i$ of $M$ is oriented toward the generated mean, $e_i^\top(\mu_g - \mu_r) \ge 0$, so $+e_i$ is the generated end.
Cohen's $d$ along a direction is the difference between the means of the generated and reference projections, divided by their pooled standard deviation.
The variance ratio is the variance of the generated projections over that of the reference projections.
A concept direction $w_c$ is the unit-normalized difference between the mean features of two groups of real samples, computed in the space that defines the metric.
The groups are reference images ranked by CLIP image-text similarity for images, and CATH domains split by class or by a geometric property for proteins.
The sparse concept fit regresses $e_i$ on the concept directions of one dictionary with a Lasso without intercept, $\min_\beta \frac{1}{2D}\|e_i - W\beta\|^2 + \alpha\|\beta\|_1$, where the columns of $W$ are the concept directions in $D$ dimensions.
It reports the surviving signed weights and the cosine between $e_i$ and its reconstruction $W\beta$.
A weight is positive when its concept grows toward $+e_i$.
We set $\alpha$ to leave a handful of concepts: $5 \times 10^{-5}$ for the CLIP dictionary, $7 \times 10^{-5}$ for the ImageNet-class dictionary, and $10^{-3}$ for the protein geometry dictionary ($10^{-5}$ in the raw space of Appendix~\ref{app:protein}).
Correlated concepts can exchange weight in such a fit.
A weight therefore need not match the alignment of its concept direction alone with $e_i$, and we name a direction by its dominant concepts rather than by every surviving weight.

\paragraph{Images.}
We generate the sets with SD1.5~\citep{rombach2022high} using the DDIM sampler~\citep{song2020denoising} at classifier-free guidance scale $7.5$ and $512 \times 512$ resolution, at $15, 20, \dots, 50$ steps, with one image per reference caption.
The COCO reference is the $30{,}000$ captioned images of the 2014 validation split.
The Inception features are the $2048$-dimensional pool features of clean-fid~\citep{parmar2022aliased}.
The CLIP dictionary uses CLIP ViT-L/14 at $336$ pixels~\citep{radford2021learning}.
It embeds each pole of a concept as the mean CLIP text embedding of its prompt set.
The two groups are the top and bottom $10\%$ of reference images by the cosine similarity of their CLIP image embedding to the difference of the two pole embeddings.
The $24$ concepts are composition simplicity, number of objects, crowd density, people, indoor/outdoor, nature versus urban, animals, food, vehicles, furniture, electronics, sports equipment, close-up versus wide shot, colorfulness, warm versus cool tone, brightness, water and sky, text and signs, staged versus candid, centered composition, depth of field, weather, daytime, and portrait versus scene.
The ImageNet-class dictionary is the $1008$ unit-normalized rows of the final Inception layer, which maps the pool features to the class logits.
The DINOv2 check uses the ViT-L/14 model~\citep{oquab2023dinov2} with the same dictionary and penalty.

\paragraph{Video.}
The reference is $2048$ UCF-101 clips of $16$ frames, one window per source video, at the $128 \times 128$ resolution both embeddings read~\citep{ge2024content}.
These $2048$ clips keep the covariance nondegenerate in the $1408$ dimensions of VideoMAE-v2.
The distortions are the elastic transformation and motion blur of ImageNet-C at its five severities.
We apply them at $128$ pixels with the suite's pixel constants, which are set for larger images, so they are harsher relative to the frame than in ImageNet-C.
FVD uses the $400$-dimensional logits of the Kinetics-400 I3D classifier.
The VideoMAE-v2 features are the $1408$-dimensional outputs of its ViT-g model fine-tuned on Something-Something-v2, following \citet{ge2024content}.
The bands for the flicker set at one severity average the displacements of the other four severities before taking eigenvectors.
The bands for the generator average all five severities.
ModelScope generates $2048$ clips of $16$ frames from the UCF-101 class names with its default DPM-Solver scheduler at $25$ steps, cropped like the reference.

\paragraph{Proteins.}
La-Proteina generates unconditional backbones from its released checkpoint.
Each setting has $467$ structures, the size of the Protein FID test set, which its default sampler spreads evenly over lengths of $100$, $200$, $300$, $400$, and $500$ residues.
The embeddings are ESM3 structure-token embeddings, mean-pooled to $1536$ dimensions with the sequence masked.
The reference is the released embeddings of $4943$ curated PDB structures of \citet{faltings2025protein}, each a FoldSeek cluster representative.
The PCA to $32$ dimensions is fit jointly on the reference and generated embeddings, as in Protein FID, and we recompute the optimal transport in this space.
The concept dictionary is built from the $31{,}885$ domains of the CATH S40 non-redundant set~\citep{sillitoe2021cath}.
These domains are embedded the same way and projected onto the same PCA basis.
The alpha-versus-beta direction runs from the mainly-alpha class to the mainly-beta class, and each geometric property is split at its terciles.
The ten geometric properties are chain length, radius of gyration, globularity (the radius of gyration over the cube root of the length), the helix, strand, and coil fractions, the number of secondary-structure segments per residue, relative contact order, clashes per residue, and backbone-bond deviation, the standard deviation of consecutive alpha-carbon spacings.

\section{Diagnostics of the image directions}
\label{app:image}

This appendix gives further diagnostics of the image directions for SD1.5 against COCO (\Secref{sec:sign_reversal}).
It separates the mean and covariance parts of their directional FD (Appendix~\ref{app:methods}) and measures the composition-simplicity concept direction on its own.

\paragraph{The top eigenvector is the mean difference.}
$e_1$ has cosine $0.99$ with the mean-difference direction at $50$ steps.
Its mean part is $85\%$ of $\mathrm{FD}(e_1)$.
The largest single share of FID is thus the mean gap between the two sets, laid along one direction.
The next eigenvectors are almost all covariance part.
Their mean part is $4\%$ of $\mathrm{FD}(e_2)$ and under $1\%$ of $\mathrm{FD}(e_3)$.
Table~\ref{tab:lasso_full} lists every concept of the sparse concept fits of Table~\ref{tab:lasso}.
It also adds the fits that each dictionary gives to the eigenvectors the other dictionary names.
The ImageNet-class dictionary reconstructs $e_1$ at cosine $0.16$, and the CLIP dictionary reconstructs $e_3$ at $0.17$.

\begin{table}[t]
  \centering
  \caption{
    Every concept whose weight rounds to at least $0.01$ in the sparse concept fits of the top eigenvectors of $M$, for SD1.5, $50$ steps, and COCO. Each weight is positive when the concept, or class logit, grows toward $+e_i$, the generated end. The last column is the reconstruction cosine.
  }
  \label{tab:lasso_full}
  \small
  \vspace{5pt}
  \begin{tabular}{@{}clp{9.6cm}c@{}}
    \toprule
    & dictionary & surviving concepts (weight) & cos \\
    \midrule
    $e_1$ & CLIP & \raggedright composition simplicity ($+0.55$), depth of field ($+0.12$), furniture ($+0.08$), electronics ($+0.05$), daytime ($+0.04$), portrait versus scene ($-0.03$) & $0.74$ \\
    $e_1$ & ImageNet & \raggedright notebook ($+0.02$) & $0.16$ \\
    $e_2$ & CLIP & \raggedright brightness ($-0.15$), weather ($-0.12$), crowd density ($+0.08$), indoor/outdoor ($+0.08$), depth of field ($+0.05$), electronics ($+0.01$), centered composition ($-0.01$), vehicles ($+0.01$) & $0.49$ \\
    $e_2$ & ImageNet & \raggedright hand blower ($+0.08$), horse cart ($-0.06$), china cabinet ($+0.06$), Crock Pot ($+0.03$), space heater ($+0.02$), crane ($-0.01$), jinrikisha ($-0.01$), tank ($-0.01$) & $0.43$ \\
    $e_3$ & CLIP & \raggedright indoor/outdoor ($+0.03$), sports equipment ($+0.01$) & $0.17$ \\
    $e_3$ & ImageNet & \raggedright Dungeness crab ($-0.09$), mashed potato ($-0.06$), burrito ($-0.05$), spiny lobster ($-0.02$), orange ($+0.02$), ice cream ($-0.02$) & $0.40$ \\
    \bottomrule
  \end{tabular}
\end{table}

\paragraph{The concept direction on its own.}
The name of $e_1$ rests on a sparse concept fit.
We therefore also measure the composition-simplicity concept direction itself, which is built without reference to $M$.
It has cosine $0.67$ with $e_1$ and carries $10.9\%$ of FID at $50$ steps.
Its directional FD has slope $0.39$ against the total FID across the step sweep of \Secref{sec:sign_reversal_sweep}, and its mean part accounts for $0.36$ of this slope.
The concept direction alone thus carries about two fifths of the FID increase.
$e_1$, whose fit it dominates, carries two thirds.

\section{Sign reversal across references}
\label{app:reversal_cumulative}

\begin{figure}[t]
  \centering
  \includegraphics[width=\linewidth]{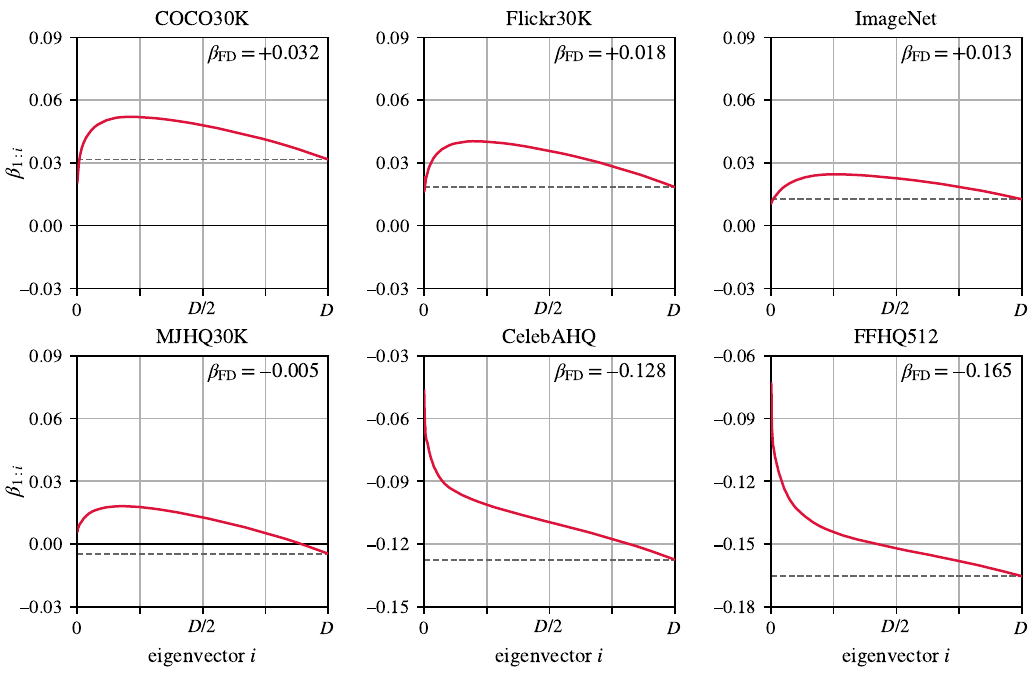}
  \caption{
    Cumulative slope $\beta_{1:i} = \sum_{j\le i}\beta_j$ of the top $i$ fixed eigenvectors against $i$. Here $\beta_i$ is the slope of the directional FD $\mathrm{FD}(e_i)$ against the step count, in each reference's own fixed $50$-step basis. The curves of COCO, Flickr, ImageNet, and MJHQ rise over the head before falling over the tail, while the face curves descend throughout. Each endpoint is the total slope $\beta_{\mathrm{FD}}$, which is positive for COCO, Flickr, and ImageNet, negative for the faces, and near zero for MJHQ.
  }
  \label{fig:reversal_cumulative_all}
\end{figure}

\Secref{sec:sign_reversal_sweep} decomposes the step-driven FID change on COCO into a rise over the head and a fall over the tail, where the rise outweighs the fall and FID rises with steps.
The sign of this change is known to differ by reference~\citep{lee2026rethinking}, so we repeat the COCO sweep on six references and ask what sets the sign.
The references are COCO30K, Flickr30K~\citep{young2014image}, ImageNet~\citep{deng2009imagenet}, MJHQ30K~\citep{li2024playground}, CelebA-HQ~\citep{karras2017progressive}, and FFHQ512~\citep{karras2019style}, spanning caption and object sets (COCO, Flickr, ImageNet), an aesthetic set (MJHQ), and two curated face sets.
Each reference is compared with SD1.5 generations from its own captions, so the six comparisons differ in the generated set as well as in the reference.
We compute each reference's own eigenbasis of $M$ at $50$ steps and hold it fixed throughout its sweep, as on COCO.
The cumulative slope $s_{1:i}$ of \Secref{sec:sign_reversal_sweep} ends at 1 on every reference, so it cannot compare the total change across references.
We instead regress each $\mathrm{FD}(e_i)$ against the step count, and the cumulative sum $\beta_{1:i} = \sum_{j \le i} \beta_j$ of the resulting slopes $\beta_i$ ends at the total slope $\beta_{\mathrm{FD}}$ of FID against steps (\Figref{fig:reversal_cumulative_all}).

\textbf{Rise and fall across references.}
The cumulative curves separate two effects that every reference shares in kind but not in size.
The first is the head, led by $e_1$, along which the generated mean drifts as steps increase.
The second is the tail, where directional FD falls with steps on every reference.
The drift along $e_1$ runs away from the reference on COCO, Flickr, ImageNet, and MJHQ, so their curves rise over the head before falling over the tail.
The drift runs toward the reference on the two face references, so their curves fall throughout, with the decrease spread across the spectrum.
The tail fall is of comparable size on every reference, so the head decides whether FID reverses.
FID rises when the head rise outweighs the tail fall, falls when the drift runs toward the reference and the head falls too, and stays nearly flat when the two cancel.

\textbf{Aligned drifts on reference-specific eigenvectors.}
The change in the generated mean across the sweep, $\mu_g(50) - \mu_g(15)$, aligns across all six comparisons (pairwise cosine $0.52$ to $0.88$), although their captions differ.
Raising the step count thus moves the generated mean in a similar direction whatever the reference.
Each reference's FID, however, is carried by its own head, so the same drift contributes differently to each FID.
The generated mean moves away from the reference along $e_1$ on COCO, Flickr, and ImageNet, where the head rise exceeds the tail fall and FID worsens.
The mean moves toward the reference along $e_1$ on CelebA-HQ and FFHQ, and directional FD also decreases broadly across their spectra, so FID improves.
MJHQ is the cancellation case: its discrepancy along $e_1$ grows like that of the caption and object references, but the tail fall offsets it and leaves the total FID nearly unchanged.

\mybox{
  \textbf{Finding.} Adding steps drives the generated mean in a similar direction in every comparison. Each reference's FID, however, is carried by its own head, so the reference decides whether FID reverses. The caption and object references (COCO, Flickr, ImageNet) worsen, the face references improve, and MJHQ cancels.
}

\section{Video bands}
\label{app:video}

\paragraph{Overlap between the bands.}
The appearance and flicker bands are computed independently from $M(R, S)$ and $M(S, T)$, so they need not be orthogonal.
The two shares of a row of Table~\ref{tab:video_bands} therefore measure how much of a distance aligns with each band, not an exclusive allocation between them.
With elastic, the two shares of I3D at $k = 10$ even sum past one.
Table~\ref{tab:video_overlap} measures this overlap at $k = 10$.
It reports the largest principal-angle cosine between the two bands and the share of their union.
It also reports the share of each band after orthogonalizing it against the other, which equals the union's share minus the other band's share.
The ordering between the two bands on the flicker set holds for both embeddings and in both orders of orthogonalization with motion blur.
The elastic bands keep the ordering too, except for I3D when the appearance band is orthogonalized against the flicker band.

\begin{table}[t]
  \centering
  \caption{
    Overlap of the $k = 10$ appearance and flicker bands. The flicker set is $M(R, T)$ at severity $3$ with bands computed from the other severities, and the generated set is $M(R, G)$ with bands computed from all severities. Columns give the largest principal-angle cosine between the bands, the share of their union, and the share of the flicker band orthogonalized against the appearance band (flicker $\mid$ appearance) and the reverse.
  }
  \label{tab:video_overlap}
  \small
  \vspace{5pt}
  \begin{tabular}{@{}llcccc@{}}
    \toprule
    embedding, family & set & cos & union & flicker $\mid$ appearance & appearance $\mid$ flicker \\
    \midrule
    I3D, motion blur & flicker set & $0.74$ & $0.746$ & $0.083$ & $0.473$ \\
    I3D, elastic & flicker set & $0.80$ & $0.833$ & $0.185$ & $0.338$ \\
    VideoMAE, motion blur & flicker set & $0.61$ & $0.786$ & $0.605$ & $0.174$ \\
    VideoMAE, elastic & flicker set & $0.69$ & $0.741$ & $0.459$ & $0.188$ \\
    \midrule
    I3D, motion blur & generated & $0.74$ & $0.384$ & $0.096$ & $0.234$ \\
    I3D, elastic & generated & $0.80$ & $0.371$ & $0.083$ & $0.265$ \\
    VideoMAE, motion blur & generated & $0.59$ & $0.183$ & $0.088$ & $0.087$ \\
    VideoMAE, elastic & generated & $0.68$ & $0.205$ & $0.056$ & $0.121$ \\
    \bottomrule
  \end{tabular}
\end{table}

\paragraph{The generator.}
The shares on the generator are smaller and closer than on the flicker set.
The two bands of VideoMAE are tied with motion blur at $k = 10$, in both orders of orthogonalization.
What VideoMAE changes relative to I3D is thus the balance between the bands.
The ratio of the flicker share to the appearance share in Table~\ref{tab:video_bands} rises from $0.52$ under I3D to $1.01$ under VideoMAE.
The elastic ratio rises from $0.37$ to $0.57$, and appearance still leads there.
The union of the two bands holds only $18\%$ to $38\%$ of the generated distance at $k = 10$.
Most of the generator's discrepancy from UCF-101 lies outside the directions that the distortions define.

\paragraph{Across severities.}
Table~\ref{tab:video_severity} compares two quantities at every severity of motion blur.
The first is the scalar ratio of \citet{ge2024content}, the flicker set's FVD over the appearance set's.
The second is the share of the flicker set's FVD on the top flicker eigenvector.
The ratio falls toward one as severity grows for both embeddings.
The share keeps the two embeddings apart at every severity.

\begin{table}[t]
  \centering
  \caption{
    Scalar ratio and top flicker-eigenvector share across the five motion blur severities, with bands computed from the other severities.
  }
  \label{tab:video_severity}
  \small
  \vspace{5pt}
  \begin{tabular}{@{}lccccc@{}}
    \toprule
    severity & $1$ & $2$ & $3$ & $4$ & $5$ \\
    \midrule
    I3D ratio & $1.41$ & $1.29$ & $1.18$ & $1.11$ & $1.07$ \\
    VideoMAE ratio & $3.56$ & $3.63$ & $3.23$ & $2.87$ & $2.77$ \\
    I3D flicker share, $k = 1$ & $0.053$ & $0.028$ & $0.023$ & $0.028$ & $0.031$ \\
    VideoMAE flicker share, $k = 1$ & $0.235$ & $0.371$ & $0.470$ & $0.475$ & $0.374$ \\
    \bottomrule
  \end{tabular}
\end{table}

\section{Protein FID in the raw embedding space}
\label{app:protein}

The main text analyzes the $32$-dimensional PCA space that defines Protein FID.
This appendix repeats the analysis in the raw $1536$-dimensional ESM3 space and lists the full sparse concept fits in both spaces.

\paragraph{The same directions.}
The raw FD is $3.40 \times 10^6$, against $2.87 \times 10^6$ for Protein FID.
The raw directional FD of the PCA-$32$ subspace is $2.90 \times 10^6$, so the lower bound of Appendix~\ref{app:methods} is nearly tight.
The raw spectrum of $M$ is less steep: $e_1$ carries $56\%$ of the raw FD, $e_2$ $12\%$, and $e_3$ $4\%$ (Table~\ref{tab:protein_full}).
The PCA thus keeps the dominant discrepancy and makes it slightly more concentrated.
The raw $e_1$ reconstructs at cosine $0.71$ and is dominated by a coil deficit.
The generated set is shifted along it by Cohen's $d = 3.2$, as in the PCA space.
The reference varies along $e_1$ more than along all but two of its principal directions in both spaces.
The lower eigenvectors keep their fits and their narrow generated spread.
The alpha-versus-beta direction carries $4.5\%$ of the raw FD with cosine $0.23$ to the mean shift, against $5.5\%$ and $0.24$ for Protein FID.

\begin{table}[t]
  \centering
  \caption{
    Full sparse concept fits of the top eigenvectors of $M$ for La-Proteina at $400$ steps, in the $32$-dimensional PCA space of Protein FID (Table~\ref{tab:protein}, penalty $10^{-3}$) and in the raw $1536$-dimensional ESM3 space (penalty $10^{-5}$). Every surviving concept is listed with its weight, and the other columns are as in Table~\ref{tab:protein}.
  }
  \label{tab:protein_full}
  \small
  \vspace{5pt}
  \setlength{\tabcolsep}{4pt}
  \begin{tabular}{@{}llcp{7.6cm}ccc@{}}
    \toprule
    space & & share & surviving concepts (weight) & cos & $d$ & var \\
    \midrule
    PCA-$32$ & $e_1$ & $66\%$ & \raggedright coil ($-0.97$), contact order ($+0.53$), length ($+0.16$), globularity ($-0.16$), radius of gyration ($-0.10$) & $0.72$ & $+3.2$ & $0.55$ \\
    & $e_2$ & $14\%$ & \raggedright backbone-bond deviation ($-0.58$), clash ($+0.50$), contact order ($+0.18$), segments per residue ($+0.17$) & $0.92$ & $+0.0$ & $0.05$ \\
    & $e_3$ & $4\%$ & \raggedright contact order ($-0.90$), clash ($+0.31$), segments per residue ($+0.18$), backbone-bond deviation ($+0.17$), radius of gyration ($+0.02$) & $0.83$ & $+0.1$ & $0.16$ \\
    \midrule
    raw & $e_1$ & $56\%$ & \raggedright coil ($-0.99$), length ($+0.51$), contact order ($+0.45$), radius of gyration ($-0.36$), alpha-versus-beta ($+0.06$), segments per residue ($-0.04$) & $0.71$ & $+3.2$ & $0.56$ \\
    & $e_2$ & $12\%$ & \raggedright backbone-bond deviation ($-0.58$), clash ($+0.50$), contact order ($+0.19$), segments per residue ($+0.18$), strand ($+0.06$), coil ($-0.06$), globularity ($+0.04$) & $0.90$ & $+0.0$ & $0.04$ \\
    & $e_3$ & $4\%$ & \raggedright contact order ($-0.82$), clash ($+0.30$), backbone-bond deviation ($+0.17$), segments per residue ($+0.13$), radius of gyration ($+0.02$), strand ($-0.01$), length ($-0.01$) & $0.76$ & $+0.1$ & $0.16$ \\
    \bottomrule
  \end{tabular}
\end{table}

\paragraph{Correlated concepts.}
The concept directions are strongly collinear in the PCA space.
The alpha-versus-beta direction has cosine $-0.98$ with the helix direction, and helix has $-0.96$ with coil.
At the raw-space penalty of $10^{-5}$, the PCA-space fit therefore trades large opposite weights among the helix, strand, and coil fractions, which sum to one.
We raise the penalty to $10^{-3}$ in the PCA space, where every weight stays below one.
Even so, a weight can differ from the alignment of its concept direction alone.
Contact order enters both fits of $e_1$ with a positive weight, although its direction alone has cosine $-0.10$ with $e_1$ in either space.
This weight is a correction among correlated concepts, not a higher contact order at the generated end.

\end{document}